\documentclass[10pt,twocolumn,letterpaper]{article}

\usepackage[pagenumbers]{cvpr} 
\usepackage[T1]{fontenc}

\usepackage{graphicx}
\usepackage{amsmath}
\usepackage{amssymb}
\usepackage{booktabs}

\usepackage[pagebackref,breaklinks,colorlinks]{hyperref}

\usepackage[capitalize]{cleveref}
\crefname{section}{Sec.}{Secs.}
\Crefname{section}{Section}{Sections}
\Crefname{table}{Table}{Tables}
\crefname{table}{Tab.}{Tabs.}

\begin{document}

\title{Speech Driven Video Editing via an Audio-Conditioned Diffusion Model}

\author{Dan Bigioi\\
University of Galway\\
{\tt\small d.bigioi1@nuigalway.ie}
\and
 Shubhajit Basak\thanks{Both Authors Contributed Equally to the Paper}\\
University of Galway\\
{\tt\small s.basak1@nuigalway.ie}
\and
Michał Stypułkowski\footnotemark[1] \\
University of Wrocław\\
{\tt\small michal.stypulkowski@cs.uni.wroc.pl}
\and
Maciej Zi\c eba\\
Wrocław University of Science and Technology\\
Tooploox\\
{\tt\small
maciej.zieba@pwr.edu.pl}
\and
 Hugh Jordan\\
Trinity College Dublin\\
{\tt\small jordanhu@tcd.ie}
\and
 Rachel McDonnell\\
Trinity College Dublin\\
{\tt\small ramcdonn@tcd.ie}
\and
 Peter Corcoran\\
University of Galway \\
{\tt\small peter.corcoran@universityofgalway.ie}
}

\maketitle

\begin{abstract}
   Taking inspiration from recent developments in visual generative tasks using diffusion models, we propose a method for end-to-end speech-driven video editing using a denoising diffusion model. Given a video of a talking person, and a separate auditory speech recording, the lip and jaw motions are re-synchronized without relying on intermediate structural representations such as facial landmarks or a 3D face model. We show this is possible by conditioning a denoising diffusion model on audio mel spectral features to generate synchronised facial motion. Proof of concept results are demonstrated on both single-speaker and multi-speaker video editing, providing a baseline model on the CREMA-D audiovisual data set. To the best of our knowledge, this is the first work to demonstrate and validate the feasibility of applying end-to-end denoising diffusion models to the task of audio-driven video editing. \footnote{All code, datasets, and models used as part of this work are made publicly available here: https://danbigioi.github.io/DiffusionVideoEditing/}
\end{abstract}

\section{Introduction}
\label{sec:intro}

The idea behind audio-driven video editing is to provide a means to re-synchronise the lip and jaw movements of an actor in a video, in response to a new speech input signal. This new speech signal may come from the original speaker, or a voice actor. Regardless of the source of the input speech, a key objective is that the performance of the actor is never diminished. No matter how the lip and jaw movements change in response to the new audio, the facial expressions, and emotions portrayed by the actor should remain consistent with the original performance. 

Achieving such seamless audio-driven video editing is an exciting prospect for the entertainment industry, one with the potential of being applied to movies, TV shows, live streaming, and even homemade content uploaded to platforms such as YouTube, TikTok, and others. Giving video content creators the ability and option to edit their work without having to go through time-consuming, and expensive re-shoots, allows them to work with a greater tolerance for error during filming.

Furthermore, the realisation of true audio-driven video editing would bring about a significant transformation in the world of cinema and television, allowing for more accessible and cost-effective dubbing of English-language movies/TV shows/videos into other languages and vice versa, allowing for the further democratisation of video content by making it more engaging and personalized for audiences worldwide. Recent advancements in deep learning and talking head generation techniques are bringing us closer to this exciting possibility, where audio and video will be seamlessly synchronized in real-time. 

Current approaches for speech driven video editing, and the related task of talking head generation can be grouped into two distinct types: structured, and unstructured. Structured generation refers to techniques that use the speech signal to first extract an intermediate structural representation of the face (facial landmarks, 3D model expression parameters), before utilizing it to render the photo-realistic frame \cite{thies2020neural, zhou2020makelttalk,ji2021audio,chen2020talking, zhao2021sparse}. On the other hand, unstructured generation techniques  \cite{vougioukas2020realistic, eskimez2020end, jamaludin2019you, zhou2021pose}, utilise image reconstruction techniques to generate the photo-realistic frame directly in an end-to-end manner. 

Diffusion models \cite{sohl2015deep, dhariwal2021diffusion, ho2020denoising, nichol2021improved} are a relatively new class of generative model that have recently been gaining traction due to their strong performance on image synthesis tasks, often outperforming state-of-the-art GAN (Generative Adversarial Network) \cite{goodfellow2020generative}-based methods \cite{dhariwal2021diffusion}. Utilising conditioning signals such as text and even images, diffusion models have shown that they can be trained and conditioned towards generating a specific desired output at inference time with relative ease \cite{rombach2022high}. They achieve high mode coverage unlike GANs, maintain high sample quality, and are stable during training. These properties make them an ideal candidate for application towards the task of unstructured audio-driven video editing, a task that has thus far been dominated by GAN-based approaches \cite{vougioukas2020realistic, chen2022talking, chen2018lip}. 

We present an approach for automatic speech driven video editing using a denoising diffusion model. We utilise a U-Net backbone modifying it for the task of video editing, and introduce a feature concatenation mechanism for conditioning the network with information related to the previously generated frame in the sequence so that the network can generate temporally coherent frames. We further condition the network on speech by feeding spectrogram feature embeddings combined with the noise signal throughout the residual layers of the U-Net as demonstrated by Diffused Heads \cite{stypulkowski2023diffused}, though unlike their approach, we use spectral features rather than features extracted from a pretrained speech encoder in order to capture as much information about the signal as possible.  To the best of our knowledge, this is the first work that applies denoising diffusion models to the task of audio-driven video editing. As part of this work, we state the following contributions to the field:

\begin{itemize}
  \item A novel unstructured end-to-end approach for audio-driven video editing using a denoising diffusion model. We condition the network on speech and train it to modify the face such that the lip and jaw movements are synchronised to the conditioning audio signal on a frame-by-frame basis. We train both single, and multi-speaker proof-of-concept models using the GRID \cite{cooke2006audio}, and CREMA-D \cite{Cao2014Crema} datasets respectively, achieving strong proof-of-concept results when tested on unseen speakers. The project code, datasets, and trained models will be made freely available to the public.  

  \item We demonstrate the applicability of our approach on the video editing task, achieving competitive results thanks to our conditional inpainting strategy which gathers information from previous frames and audio spectral embeddings, to generate the current frame. Our method achieves near state-of-the-art results when measured on traditional image quality metrics such as SSIM, PSNR, FID, and competitive SyncNet \cite{Chung16a} lip synchronisation scores compared to other relevant methods from the field. 

\end{itemize}

\section{Related Works}
\label{sec:related_works}

\subsection{Audio Driven Video Generation}

Audio-driven video generation methods can generally be categorised by whether they are generated by leveraging an audio-driven structural representation of the face, or without. 

There have been numerous approaches over the years relating to the former. Taylor et al. \cite{taylor2017deep} and Karras et al. \cite{karras2017audio} among the first to apply machine learning techniques to the facial animation task, the former learning facial expression parameters of a 3D face model from phoneme labels, and the latter predicting 3D vertex positions of a face mesh from a speech audio window. Suwajanakorn et al. \cite{suwajanakorn2017synthesizing} trained a speaker specific network to output sparse mouth key-points, using them to modify videos of President Obama. Eskimez et al \cite{eskimez2018generating} presented a recurrent architecture capable of taking in speech as input and outputting 2D landmark face co-ordinates, with Chen et al. \cite{chen2019hierarchical} utilising cascaded GANs to translate those landmark features into photorealistic frames. Cudeiro et al. \cite{cudeiro2019capture} introduced a 4D facial dataset, and trained a network to generate animations from speech with it. \cite{zhou2020makelttalk, wang2021audio2head, lu2021live, biswas2021realistic} generated intermediate landmark features from audio, also exploring the related task of extracting realistic headpose. Thies et al. \cite{thies2020neural} generated 3D facial expression parameters using features from a pretrained audio encoder, using these parameters to generate a photorealistic video via a neural renderer, with \cite{wen2020photorealistic} and \cite{song2022everybody} following a similar approach but operating on videos instead. \cite{chen2020talking} and \cite{yi2020audio} presented methods to generate 3D face animation parameters, in addition to realistic head pose from speech, using these features to generate photorealistic frames. Ji et al. \cite{ji2021audio} approached the problem of video editing, generating emotion-controllable talking head portraits using both intermediate landmark structures, and 3D model parameters. \cite{zhang2021facial, zhang2021flow,wu2021imitating, lahiri2021lipsync3d, richard2021meshtalk, song2021tacr} are other approaches from the literature which predict expression parameters from audio to drive a 3D face model. 

What these approaches all have in common is that they use these intermediate structural representations as input to a separate neural rendering model which is typically trained as an image-to-image translation task to generate the final photo-realistic image frame. As of the date of this submission, GAN-based \cite{goodfellow2020generative} approaches such as Pix2Pix \cite{isola2017image}, CycleGAN \cite{zhu2017unpaired}, and other variations have proved immensely popular for this task. However, diffusion-based techniques show big promise for the future, especially given recent developments in various image-to-image translation tasks \cite{saharia2022palette}. 

Nonstructural/end-to-end methods on the other hand utilise latent feature learning and image reconstruction techniques to generate a photo-realistic video sequence from an input speech signal and reference image/video in an end-to-end manner. Approaches such as \cite{prajwal2020lip, mittal2020animating, zhu2021arbitrary, eskimez2020end, chen2018lip, zhou2021pose, vougioukas2020realistic, kumar2020robust, zhou2019talking, song2019talking, jamaludin2019you} have seen much success in recent times. Each of these approaches differs from the one used in this paper as they are all GAN/VAE (variational autoencoder)\cite{kingma2013auto} based probabilistic methods while ours leverages a denoising diffusion model. While current end-to-end approaches suffer from low output resolution quality compared to structural methods, there is a lot of potential for improvement, especially by exploiting diffusion models' ability to synthesise high-quality samples while maintaining good mode coverage/diversity.  

\subsection{Diffusion Models}

Denoising diffusion models \cite{sohl2015deep, song2019generative} have seen great success on a wide variety of different challenges, ranging from image-to-image translation tasks like inpainting, colorisation, image upscaling, uncropping \cite{saharia2022palette, ho2022cascaded, batzolis2021conditional, preechakul2022diffusion, rombach2022high, saharia2022image, lugmayr2022repaint, meng2021sdedit}, audio generation \cite{chen2020wavegrad, kong2020diffwave, yang2022diffsound, tae2021editts, popov2021grad, kim2022guided, huang2022prodiff, levkovitch2022zero}, text-based image generation \cite{avrahami2022blended, ruiz2022dreambooth, fan2022frido, nichol2021glide, ramesh2022hierarchical, saharia2022photorealistic, gu2022vector}, video generation \cite{ho2022video, zhang2022motiondiffuse, harvey2022flexible, yang2022VIDEO}, and many others. Recently, diffusion models have also been applied to the related task of talking head generation, with the work of \cite{stypulkowski2023diffused} a concurrent approach to our own. For a thorough review of diffusion models and all of their recent applications, we recommend \cite{yang2022diffusion}. 

Diffusion models are a class of generative probabilistic models that consist of two steps: 1) the forward diffusion process that destroys data by steadily adding small amounts of random Gaussian noise over a series of time steps until the data becomes a sample from a standard Gaussian distribution. 2) The reverse diffusion process where a denoising model is trained to restore structure in the data by steadily removing noise over a series of time steps. The trained model can then sample information from random Gaussian noise and steadily denoise it over a series of time steps to attain the desired output. 

Sohl-Dickstein et al.~\cite{sohl2015deep} developed the first diffusion model and coined the term, followed by Ho et al.~\cite{ho2020denoising} combining denoising score matching with Langevin dynamics \cite{song2019generative} and diffusion models to synthesise images. This ignited a steady interest in diffusion models, with Nichol et al.~\cite{nichol2021improved} showing that by making small adjustments to the diffusion process, they could sample data faster and achieve better log-likelihoods to models trained explicitly to minimise it with minimal impact to sample quality. They also found that training diffusion models with more computational power typically lead to better sample quality. Chen et al.~\cite{chen2020wavegrad} and Kong et al.~\cite{kong2020diffwave} applied diffusion models to the task of audio synthesis, succeeding in generating high-quality samples. Dhariwal and Nichol~\cite{dhariwal2021diffusion} demonstrated that diffusion models beat GANs on image synthesis, also introducing the concept of "classifier guidance" for a conditional generation. 

As diffusion models are trained under a single loss and do not rely on a discriminator, they are more stable during training and do not suffer from typical issues associated with training GANs such as mode collapse, and vanishing gradients. They produce high-quality output samples and display high mode coverage unlike GANs \cite{xiao2021tackling}. Despite these advantages, their sampling speed is slow due to the need to run the inverse diffusion process thousands of times on the same sample to denoise it completely. Xiao et al.~\cite{xiao2021tackling} and Rombach et al~\cite{rombach2022high} attempted at speeding up the sampling and training times associated with diffusion models with the former proposing a method to model the denoising distribution using a complex multi-modal distribution in order to facilitate larger diffusion steps, and the latter applying diffusion models in the latent space of a pre-trained autoencoder to reduce the complexity. This is an ongoing focus of research in the field, and it is a certainty that more works tackling the inference/training speed problem will emerge. 

\section{Methodology}

A diffusion model is defined as having two steps, the forward diffusion process where the data is gradually destroyed, and the learned inverse diffusion process which reconstructs the data, and is used during training and inference. In our case, we condition a denoising U-Net on image and speech features to denoise a masked portion of the target frame into the desired output. A high-level overview of this process is depicted in figure \ref{fig:network_architecture}. 

\subsection{Diffusion process}

\subsubsection{Forward diffusion process}

As defined by \cite{sohl2015deep}, the forward diffusion process is a Markov chain that adds small amounts of noise to the data $y$ over a predefined number of time steps T, until the data is completely destroyed at time step t=T. This state is represented as $y_T$ with $y_0$ representing the data before any noise was added to it. The Markov chain is defined by:   

\begin{equation}
  q\left ( y_{1:T}|y_{0}\right ):=\prod_{y=1}^{T}q\left ( y_{t}|y_{t-1}\right )
  \label{eq:Markov Chain}
\end{equation}
where at each step, Gaussian noise is added by:

\begin{equation}
  q\left ( y_{t} | y_{t-1} \right  ) := \mathcal{N}\left ( y_{t}; \sqrt{\alpha_{t}}y_{t-1},(1-\alpha_{t})I \right ),
  \label{eq:Gaussian_Noise_Step}
\end{equation}
with $\alpha_{t} := (1 - \beta_{t})$, representing the hyperparameters of our fixed noise scheduler. \cite{ho2020denoising} show that it is possible to sample $y_t$ at any step $t$ in closed form: 

\begin{equation}
  q\left ( y_{t} | y_{0} \right  ) := \mathcal{N}\left ( y_{t}; \sqrt{\bar{\alpha_{t}}}y_{0},(1- \bar{\alpha_{t}})I \right ),
  \label{eq:Gaussian_Noise_Step_closed_form}
\end{equation}
with $\bar{\alpha_{t}} := \prod_{s=1}^{t}\alpha_s $. This is an important observation, as it significantly speeds up the forward diffusion process, and can be used to train a model on the fly with random noise levels at each forward step. 

\subsubsection{Inverse diffusion process}

Given a noisy image $\bar{y}$ defined as: 

\begin{equation}
  \bar{y} :=  \sqrt{\bar{\alpha}}y_{0} + \sqrt{1 - \bar{\alpha}}\epsilon, \epsilon \sim  \mathcal{N}(0,I)
  \label{eq:noisyImage}
\end{equation}
the goal of the Inverse diffusion process is to learn an algorithm that can denoise and restore the noisy image to its original image $Y_0$. Following the approach in \cite{saharia2022palette}, we train a neural network $f_{\theta}(x,\bar{y},\bar{\alpha}, \omega)$, a 2D U-Net\cite{ronneberger2015u}, to predict the noise generated at time $t$, optimising the $L_{simple}$ objective proposed by \cite{ho2020denoising}:
\begin{equation}
 \mathbb{E}_{t,y{_0},\epsilon}\bigg[\left\|f_{\theta}(x, \sqrt{\bar{\alpha}}y_{0} + \sqrt{1 - \bar{\alpha}}\epsilon, \bar{\alpha}, \omega) - \epsilon \right\|^2\bigg]
  \label{eq:Lsimple}
\end{equation}
where $x$ represents the identity and previous frame input to our network, $\bar{y}$ the noisy image, $\bar{\alpha}$ the noise level, and $\omega$ the audio features. During training, we only calculate the loss for the masked region of the face to conserve computational resources, following the approach in \cite{saharia2022palette}.

Following \cite{ho2020denoising}, to run inference, each step of the inverse diffusion process can then be computed by:
\begin{equation}
    y_{t-1} \leftarrow \frac{1}{\sqrt{\alpha_{t}}} \left ( y_{t} - \frac{1 - \alpha_{t}}{\sqrt{1-\bar{\alpha_{t}}}} f_{\theta}(x,y_{t}, \bar{\alpha_{t}}) \right  ) + \sqrt{1-\alpha_{t}}\epsilon_{t},
  \label{eq:backwards_diffusion_step}
\end{equation}
where $\epsilon \sim  \mathcal{N}(0,I)$. The inverse diffusion step is then repeated T times. Please see figure \ref{fig:network_architecture}
for a high-level view of our network architecture, and to better understand where each equation is used. For a more detailed discussion behind these equations, and how they are derived, please see \cite{ho2020denoising,sohl2015deep, song2019generative}. 

\subsection{Model Architecture}
\label{sec:model_architecture}

\begin{figure*}[t]
  \centering
  \includegraphics[width=\linewidth]{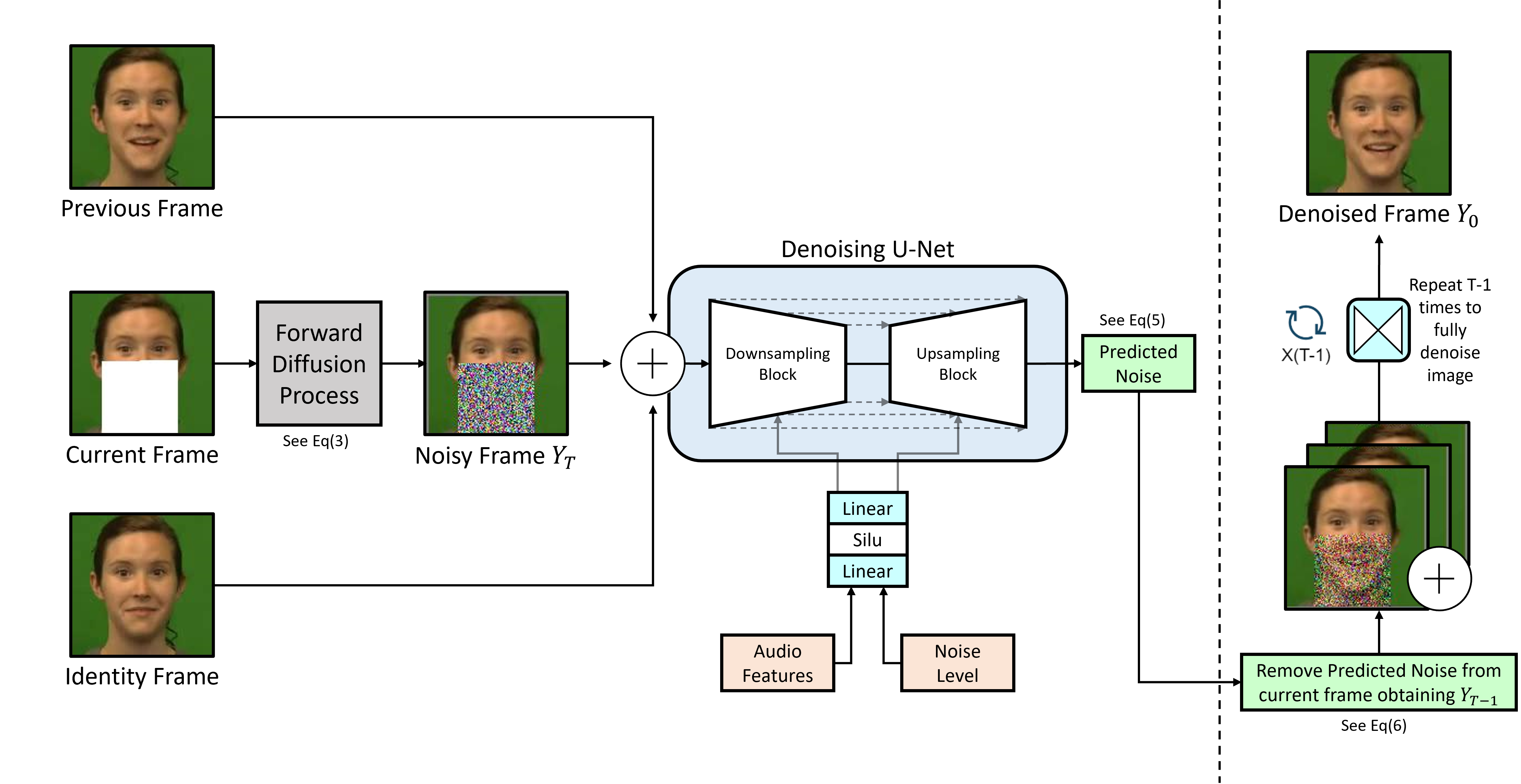}
  
   \caption{High-level overview of the network architecture. Left of the dashed line indicates the training procedure, right of it depicts the inference procedure. $\bigoplus$ represents the concatenation operator, and $\dashrightarrow$ represents a skip connection. The current frame is passed through the forward diffusion process where the noise is computed and added to the masked region of the face, obtaining noisy frame $Y_{t}$ (Equation \ref{eq:Gaussian_Noise_Step_closed_form}). The previous and identity frames are then concatenated channel-wise to it, forming a 128x128x9 feature and passed to the U-net directly. Audio features and noise level information are fed into the U-net through conditional residual blocks as described in equation \ref{eq:audio_time_conditioning}. During inference, the predicted noise is removed from noisy image $Y_{t}$, obtaining $Y_{t-1}$. The previous and identity frames are concatenated to $Y_{t-1}$, and the process is repeated until the image is fully denoised (Equation \ref{eq:backwards_diffusion_step}). 
   \label{fig:network_architecture}
   }
\end{figure*}

Figure \ref{fig:network_architecture} depicts the overall architecture of our model. We frame the problem of audio-driven video editing as a conditional inpainting task with a few key changes. Traditionally, inpainting is an image-to-image translation task where a neural network must learn to fill in a masked out region of the image with realistic content. For video editing, we must provide the network with additional context, to help guide its generation process. To do this, we split the conditioning step into two categories, frame-based, and audio-based conditioning. 

\textbf{Frame-Based Conditioning:} For a given frame $y^{i}$ extracted from a video consisting of frames ($y^{0}$,...,$y^{n}$), our model takes three images as input: 1) the current \textit{masked noisy frame} $y^{i}_{T}$ that is to be inpainted, 2) the \textit{previous frame} $y^{(i-1)}$ in the video sequence, and 3) a constant \textit{identity frame} $y^{0}$. As our approach is auto-regressive and works on a frame-by-frame basis, the purpose of the previous frame is to ensure that there is temporal stability between consecutive frames. Omitting it causes the model to output jittery, unstable frames. The identity frame is there to encourage the model not to deviate away from the target identity during the generation process, as so often is the case with auto-regressive models. While the identity frame can be omitted if training a single-speaker model with little to no adverse effects, we found that its inclusion was key to having a model that could generalise well to unseen subjects when training on multiple identities. These three frames are concatenated channel-wise, and fed into the U-Net as an input feature of size [128x128x9], as depicted on the left hand side of figure \ref{fig:network_architecture}.

\textbf{Audio-Based Conditioning:} 

For a given video sequence of frames ($y^{0}$ ,..., $y^{n}$), there is a corresponding sequence of audio spectral features ($spec^{0}$ ,..., $spec^{2n}$) extracted from the original speech signal. Each audio feature spans a 40ms window, overlapping every 20ms. Details on how we compute these features are provided in section \ref{subsec:dataprocessing}. In order to provide the audio information to the network, we extract a window of audio from ($spec^{2i-2}$ to $spec^{2i+2}$) spanning 120 ms denoted as $z^{i}$ that is centered around the current video frame $y^{i}$. We do this so that audio information from both the preceding and following frames is captured within the window to guarantee the accurate production of lip movements for plosive sounds ("p, t, k, b, d, g") by taking into consideration that these lip movements precede the sound production. We introduce this information to the U-net via the use of conditional residual blocks that condition the network on audio and noise level embeddings, scaling and shifting the hidden states of the U-net following the approach of \cite{stypulkowski2023diffused}: 

\begin{equation}
    h_{s+1} = z^{i}_{s} (t_{s}GN(h_{s}+t_{b}))+z^{i}_{b}
  \label{eq:audio_time_conditioning}
\end{equation}

where $h_{s}$ and $h_{s+1}$ represent consecutive hidden states of the U-Net,  ($z^{i}_{s}, z^{i}_{b}$) = MLP$(z^{i})$, and ($t_{s}, t_{b}$) = MLP$(\bar{\alpha}_{t})$. MLP represents a shallow neural network with a couple of linear layers separated by a SiLu() activation function, and GN is a group normalisation layer. This is shown in figure \ref{fig:network_architecture}. 

\textbf{U-Net Set Up: }In order to denoise the current noisy frame, we use a denoising U-net \cite{ronneberger2015u}, following the general architecture described by \cite{saharia2022palette}, which in turn is based on the model proposed by \cite{ho2020denoising} with modifications inspired by the works of \cite{dhariwal2021diffusion, saharia2022image}. For this work we use a lightweight 128x128 version of the 256x256 U-net architecture described by \cite{dhariwal2021diffusion}, omitting the class conditioning mechanism. Like \cite{saharia2022palette} we condition the model to generate the desired frames via the concatenation of the previous and identity frames to the masked frame. We drive the facial animation by sending audio features throughout conditional residual blocks within the U-Net as detailed by \cite{stypulkowski2023diffused}, described by equation \ref{eq:audio_time_conditioning}. We include all details related to our U-Net configuration in table \ref{table:hyperparameters}. 

Table \ref{table:hyperparameters} displays the hyperparameters we use to train our diffusion model for the task of audio-driven video editing. We train two models, a single-speaker model trained on identity S1 of the GRID dataset, and a multispeaker model trained on the train set of the CREMA-D dataset. A notable difference between the two models is the use of attention. For the single-speaker model, we omitted it from the up/downsampling layers of the U-Net, using it only within the middle block in an effort to boost training speed. Despite this, we still obtain pleasing results, as shown both in table \ref{tab:results_table}, and in the videos provided as part of the supplementary materials. During our experiments, we discovered that the use of attention within the multi-speaker model was crucial for it to generalise well to both seen and unseen speakers. We apply it at resolutions of 32x32 within the up/downsampling layers of the U-Net. We provide more discussion on this in section \ref{sec:experiments_and_results}. To perform training we used a server of 4 32GB V100 GPUs, allowing us a batch size of 40 per GPU.

\begin{table}
  \centering
  \begin{tabular}{ccc}
    \toprule
     & Single ID & Multi-ID\\
    \midrule
    Image Size & $128x128$ & $128x128$\\
    Total Frames & $73704$ & $432000$\\
    Diffusion Steps & $2000$ & $1000$\\
    Noise Schedule & $Linear$ & $Cosine$\\
    Linear Start & $1e-06$ & $NA$ \\
    Linear End & $0.01$ & $NA$ \\
    Input Channels & $10$ & $10$\\
    Inner Channels & $64$ & $64$\\
    Channels Multiple & $1,2,4,8$ & $1,2,3$\\
    Attention Resolution & $NA$ & $32$\\
    Res Blocks & $2$ & $2$ \\
    Head Channels & $32$ & $32$\\
    Drop Out & $0.2$ & $0.2$\\ 
    Batch Size & $10$ & $40$\\
    Training Epochs & $2000$ & $735$\\
    Learning Rate & $5e-05$ & $5e-05$\\
    \bottomrule
  \end{tabular}
  \caption{U-Net Training Hyperparameters}
  \label{table:hyperparameters}
\end{table}

\subsection{Data Processing}
\label{subsec:dataprocessing}

\subsubsection{Dataset}
 We rely on the GRID \cite{cooke2006audio}, and CREMA-D \cite{Cao2014Crema} audio-visual speech data sets to carry out the work in this paper. GRID is a multi-speaker data set consisting of 34 speakers (18 male, 16 female), with each speaker uttering 1000 short 6-word sentences. CREMA-D is a multi-speaker dataset consisting of 7,442 talking head clips of 91 speakers from diverse ethnic backgrounds. We present two models: 1) A single speaker model trained on 950 videos from the speaker 1 of the GRID dataset, with the model's performance being evaluated on the remaining 50 videos on the task of video editing. 2) A multi-speaker model trained on a majority of the CREMA-D dataset, with videos from identities 15, 20, 21, 30, 33, 52, 62, 81, 82, and 89 kept hidden from the model for testing and evaluation purposes. 

\subsubsection{Audio Preprocessing}

 From each video within the GRID and CREMA-D datasets, we extract the audio files and resample them at 16Khz. From the audio we compute overlapping mel-spectrogram features with n-fft 2048, window length 640, hop length 320, and 256 mel bands. With these values, a 1-second audio feature has a shape [50,256] that can be easily aligned to a sequence of video frames.
 
\subsubsection{Video Preprocessing}\label{video_processing}

First, we perform a 128x128 pixel crop centered around the face on every video frame. We do this by aligning the face in the video to the canonical face with a smoothing window of 7 frames, following the approach of \cite{Vougioukas2018EndtoEndSF}. We do this for two reasons: To get rid of any irrelevant background, and to reduce the image size to facilitate faster training and convergence speeds. In our initial experiments, we used an image size of 256x256 however the model was too expensive to train on our limited resources. It is worth noting that a video super-resolution technique such as \cite{liu2022learning} may be applied on top of our solution to achieve high-resolution samples.

Next, every video frame needs to have a rectangular region of the face masked out. Using an off-the-shelf facial landmark extractor \cite{lugaresi2019mediapipe}, we extract facial landmark coordinates to determine the position of the jaw. Using this information, we mask out a rectangular portion of the face that covers a region just below the nose, as within figure \ref{fig:network_architecture}. This face mask is computed and applied to the frames at train time within the data loader on the fly. 

During training, it is critical to hide the speaker's jawline with a rectangular face mask. This is because the network can easily pick up on the strong correlation between lip and jaw movements, leading it to ignore the speech input entirely. By hiding the jawline, we compel the model to learn to generate lip movements based solely on the accompanying speech. As the diffusion process relies on a single loss function, applying the rectangular face-mask is the easiest way to prevent the frame-based input dominating over the speech input. 

\subsubsection{Audio Video Alignment:}

As described previously in section \ref{sec:model_architecture}, given a video sequence with frames ($y^{0}$ ,..., $y^{n}$), there is a corresponding sequence of audio spectral features ($spec^{0}$ ,..., $spec^{2n}$) extracted from the original speech signal. Each audio feature spans a 40ms window, overlapping every 20ms. For any given frame $Y^{i}$, it is aligned to audio features spanned by ($spec^{2i-2}$ to $spec^{2i+2}$). To align the first and last video frames, we simply append silence to the start, and ends of their respective audio features. Care must be taken when choosing the audio window, too large and the network won't use the most meaningful information available to it, too small and there may not be enough context for the network to generate more complex lip movements caused by plosives. 

\section{Experiments \& Results}
\label{sec:experiments_and_results}

\begin{table*}[t]
\centering
\small

\begin{tabular}{ccccccccc}
\toprule
Method & LSE-C$\uparrow$ & LSE-D$\downarrow$ & FID & SSIM$\uparrow$ & PSNR$\uparrow$ & CPBD\\
\midrule
Ground Truth CREMA-D  & 5.45 & 8.12 & - & - & - & - \\
EAMM (Actual) & 3.98 & 8.92 & 22.52  & 0.74 & 29.43 & 0.1 \\
EAMM (Random) & 3.95 & 8.98 & 23.04 & 0.72 & 29.21 & 0.124 \\
PC-AVS (Actual) & 6.12 & 7.8 & 38.46  & 0.61 & 28.47 & 0.127 \\
PC-AVS (Randon) & 6.07 & 7.82 & 40.05 & 0.59 & 28.42 & 0.11 \\
SpeechDrivenAnimation & - & - & 155.63 & $0.844^*$ & $27.98^*$ & $0.277^*$ \\
Wav2Lip(Actual) & 5.89 & 7.57 & 16.21 & 0.886 & 34.23 & 0.253 \\
Wav2Lip(random) & 5.6 & 7.89 & 20.23 & 0.872 & 34.04 & 0.247 \\
Make It Talk & 3.5 & 9.71 & 27.35 & 0.75 & 31.37 & 0.152 \\
Ours (MultiSpeaker - 100) & 3.53 & 9.74 & $2.362^\dagger$ & 0.893 & 34.32 & 0.26  \\
Ours (MultiSpeaker - 500) & 3.5 & 9.68 & $2.13^\dagger$ & 0.902 & 34.4 & 0.26  \\
Ours (MultiSpeaker - 1000) & 3.49 & 9.69 & $2.369^\dagger$ & 0.863 & 34.12 & 0.242 \\
Ours (Single Speaker) & 4.98 & 7.59 & $2.312^\dagger$ & 0.92 & 32.47 & 0.29 \\
\bottomrule
\end{tabular}

\caption{
Quantitative comparison with previous works on image quality and lip synchronization metrics. Most previous works we compare to require a driving video to guide the pose of the generated speaker. For these approaches (Actual) indicates whether we provided the ground truth video to their model in addition to the ground truth audio to generate the new video, while (Random) indicates that we used a random audio file instead. We report their results under both configurations to maintain fairness. For our models we also indicate how many diffusion timesteps were used to generate the frames during inference. We report results for 100, 500, and 1000 inference steps. $^\dagger$ indicates that this metric was computed on the full frame. $^*$ indicates that these results are reported from their paper. }
\vspace{-0.05in}
\label{tab:results_table}
\end{table*}

In this section we present two models. A single-speaker video editing model trained on speaker S1 from the GRID dataset, and a multi-speaker model trained on the train-split of the CREMA-D dataset. We evaluate and compare our results to other recent audio-driven video generation methods, namely EAMM \cite{ji2022eamm}, PC-AVS \cite{zhou2021pose}, MakeItTalk \cite{zhou2020makelttalk}, Speech Driven Animation \cite{vougioukas2020realistic}, and Wav2Lip \cite{prajwal2020lip}. All models we test against are relevant end-to-end image-reconstruction based methods, except for MakeItTalk, a landmark-based method we compare against for reference purposes. We evaluate these models on the CREMA-D multispeaker test set, reporting their scores along with our own in table \ref{tab:results_table}. We generate the videos for each model using the official publicly available implementations with the recommended parameters. 

As our models are trained explicitly for video editing, they generate only a small portion of the overall frame, while keeping the rest as is. Therefore, to maintain fairness, all metrics that rely on comparing the generated frame to the ground truth are computed only on the generated portion of the image. This limitation could also create bias in the perceptual metrics and readers should consider this when comparing our model scores to others within the literature. 

We emphasise that the objective of this paper is to serve as a proof-of-concept demonstrating the potential of applying denoising diffusion models to the task of audio-driven video editing. As such, while we do not achieve state-of-the-art in some of the metrics we report, our results still show promising improvements over existing methods and highlight the potential of using denoising diffusion models for this task instead of traditional GAN-based methods.



\subsection{Evaluation Metrics}

\begin{figure}[t]
  \centering
  \includegraphics[width=\linewidth]{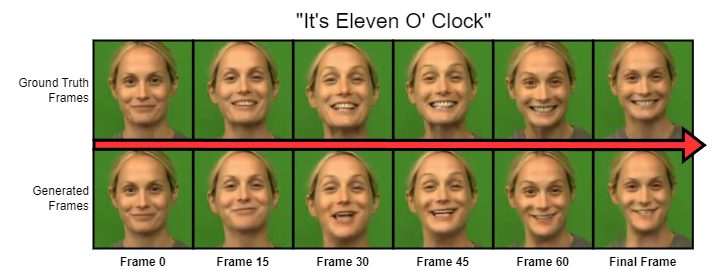}
   \caption{Multi-speaker failure case: Over time the appearance of the speaker slowly drifts away from the original.}
   \label{fig:failcase}
\end{figure}

We use a number of objective metrics to measure the quality of our generated videos, allowing us to compare them directly to other state-of-the-art audio-driven video generation methods from the literature. We compute SSIM \cite{wang2004image} (Structural Similarity Index Measure) $\uparrow$, PSNR (Peak Signal to Noise Ratio) $\uparrow$, and FID \cite{heusel2017gans} (Frechet Inception Distance) $\downarrow$ scores for the generated videos against their corresponding ground truth to measure the overall quality of the generated frames. We also compute CPBD \cite{narvekar2011no} (Cumulative Probability Blur Detection) $\uparrow$ scores, and SyncNet \cite{prajwal2020lip, Chung16a} Confidence (LSE-C) $\uparrow$ and Distance (LSE-D) $\downarrow$ scores. We reiterate the point that in order to maintain fairness when computing the image quality metrics, we only compute them on the generated portion of the image where possible. 

\subsection{Single Speaker}\label{single_speaker}

We train our single speaker model on identity S1 using data from the GRID audio-visual corpus \cite{cooke2006audio}. There are 1000 videos in total, each of them roughly 3 seconds in length totaling about 50 minutes of audio-visual content for training. We train our model on 950 videos, withholding 50 of them for testing purposes. We train this model for 895 Epochs. As we mentioned previously, we did not use any attention layers within the up/downsampling blocks of this model, using it just within the middle block of the U-Net. We did this to save on training time, however, for stronger results we recommend using it, as we show within our multi-speaker model.

\subsection{Multi-Speaker} 

We train our multi-speaker model on all identities of the CREMA-D data set except for speakers 5, 20, 21, 30, 33, 52, 62, 81, 82, and 89, choosing to keep them hidden from the model for testing purposes. We train the model for 735 Epochs. There are a number of key changes we make to train the multi-speaker model. First, we use self-attention layers within the U-Net at the 32x32 resolution, as well as in the middle block. Second, we switch to a cosine noise schedule and decrease the number of diffusion steps taken by the model during training to 1000. Finally, we decrease the number of channel multiples to [1,2,3]. We also experimented with training a model without attention in the up/downsampling blocks. It failed to converge on even train set identities. We speculate that increasing the number of inner channels used by our U-Net from 64 to 128 or 256 would significantly improve the results, as well as training the model for a longer amount of time.  Please see table \ref{tab:results_table} for a summary of our experiments and evaluations, compared to other popular works in the literature, and section \ref{subsec:results_discussion} for a detailed discussion surrounding the results.

\subsection{Results Discussion} 
\label{subsec:results_discussion}

Table \ref{tab:results_table} depicts the results our models score when tested on their unseen test sets versus other approaches in the literature. While the results we obtain are not state-of-the-art in all metrics, they successfully demonstrate that using a denoising diffusion model to do audio-driven video editing, is indeed quite feasible, and produces high-quality results comparable to other relevant methods in the literature. 

The multi-speaker model generalises quite well to unseen speakers, scoring highly on image quality metrics, managing to outperform all other methods except for Wav2Lip on SSIM and CPBD. The single speaker model also achieving similarly strong results. We believe that this is due to the diffusion models inherent ability to model complex, high-dimensional data distributions, allowing it to learn the statistical properties of the dataset and generate images that are similar to those in the training set. Further, as diffusion models are trained to gradually remove noise from the target image over time, this may help it generate smoother, and more visually pleasing results than those generated by a GAN-based model which generates the frame in one shot. Within the context of audio-driven video editing, achieving visually pleasing results is a key requirement that our model fulfills. Please see the videos attached in the supplementary material for a visual comparison between our method and existing ones. 

When evaluated on SyncNet \cite{Chung16a} confidence (LSE-C) and distance (LSE-D) scores, our multi-speaker results are comparable to other popular methods from the literature, slightly outperforming MakeItTalk, but scoring lower than EAMM. PC-AVS and Wav2Lip score the highest in that order. Notably, their approaches significantly outperform the ground truth. We believe that this is because all other methods are specifically trained to optimise a loss function designed to penalise their models for poor lip synchronisation. In the case of PC-AVS and Wav2Lip, they both rely on a strong lip sync discriminator, to encourage their models to generate distinct, clear lip movements given speech. Our approach uses no such losses or discriminators, inherently learning the relationship between speech and lip movement during training. As such while our lip synchronisation scores on unseen speakers are lower, we offer a novel approach to the task as we do not explicitly train the model to improve lip synchronisation. 

It is also worth noting that our single-speaker model performs very well on the synchronisation metrics mentioned above, leading us to speculate that with more time spent learning the data distribution, our multi-speaker model could also theoretically achieve such results.  

During inference, we noticed that the multi-speaker model occasionally struggled to maintain the identity consistent throughout the generation process, with the problem especially prevalent if there were extreme changes in head pose present in the original video. This is due to a buildup of small errors, as our approach is completely auto-regressive at inference time, relying entirely on just the previously generated frame, and identity frame to modify the current frame. Figure \ref{fig:failcase} highlights one such instance of failure, and the phenomenon is noticeable in some of the videos we provide in our video abstract. We speculate that this could be alleviated in three ways 1) introducing small amounts of face warping on the previous frame during training in order to simulate the distortion that naturally occurs over the generation process. This would encourage the model to look at the identity frame in order to correct itself. 2) Simply train the model for longer. 3) Train on a more diverse dataset of speakers captured in unconstrained conditions such as VoxCeleb or LRS. 

When testing on identities seen by the network during training by replacing the original audio with a new one, the model achieves strong lip synchronisation, and the identity deviation seen when testing on unseen identities is significantly diminished, or simply does not occur over the course of the video. This problem is also non-existent in our single-speaker model.  

We also observed that the multi-speaker model is highly sensitive to speaker volume, and intonation, especially when exposed to speech from unseen speakers. In instances where the speaker shouts or speaks loudly and clearly at the microphone, the lip movement is highly accurate and appears well-synchronised. When the volume is low, the speaker appears to be mumbling, and the full range of lip motion is not correctly generated. Analysing the synchronisation metrics confirmed this for us, with videos generated using audio labelled as being "angry" or "happy", scoring significantly higher than instances where the portrayed emotion was "sad", "fearful", or "disappointed". We suspect that this is due to our use of spectral feature embeddings when conditioning our network, and could be alleviated or significantly diminished with the use of a pretrained audio encoder for speech recognition. This is because such models are typically trained to extract the content from speech, disregarding information considered irrelevant such as pitch, or tone, and intonation.

\section{Future Work}
\textbf{Model Speed and In The Wild Training:} It is no secret that diffusion models are slow, both to train and to sample from. Our models are no exception, taking us approximately 6 minutes/epoch to train the single-speaker model, and 40 minutes/epoch for the multi-speaker one. We briefly experimented with training in the latent space to speed up training following the approach of \cite{rombach2022high}, however, sample quality suffered, so we decided to operate in the pixel space. We intend to revisit this however as improving our models training speed is a top priority for us as it would allow us to train on larger, more diverse, "in-the-wild" datasets such as VoxCeleb \cite{nagrani2017voxceleb}, or LRS \cite{chung2017lip}.  

\textbf{Appearance Consistency}: As previously discussed, our multi-speaker model's generated output appearance for unseen identities occasionally deviates from the original. To investigate this phenomenon, we intend to delve deeper into the underlying causes. Specifically, we will explore whether this effect is due to inadequate training or insufficient diversity in the training dataset, or a combination of both. By conducting a more detailed analysis, we hope to gain a better understanding of how to optimize our model's performance for a wider range of identities. Further, we intend to fully train a model that utilises the face warping augmentation to determine whether this truly provides a positive impact on the generated samples. 

\textbf{Speech Conditioning}: We plan to explore the potential of conditioning our model with a wider range of speech features, such as experimenting with larger or smaller window sizes when computing spectral features or using pre-trained audio encoders such as Wav2Vec2 \cite{baevski2020wav2vec}, Whisper \cite{radford2022robust}, or DeepSpeech2 \cite{amodei2016deep}. We believe that incorporating such features could potentially improve the lip synchronization performance of our model and generate even more realistic and expressive lip movements.

\section{Conclusion}

Our results showcase the versatility of denoising diffusion models in capturing complex relationships between audio and video signals and generating coherent video sequences with accurate lip movements for the task of speech-driven video editing. We are encouraged by the strong performance achieved by our proof-of-concept approach, scoring highly on all tested metrics, comparable to existing state of the art in end-to-end video generation. 

However, our work is not without limitations. The CREMA-D dataset is relatively small compared to other publicly available speech and video datasets, which limits the generalizability of our approach to other domains. Additionally, our approach requires a significant amount of computational resources and time to train. This is a challenge for real-time applications or for training on large-scale datasets.

We are confident that our work will inspire further research and development in this area, leading to more efficient and effective methods for speech-driven video editing. With the continuing advancements in machine learning and computer vision, we believe that denoising diffusion models will play an increasingly important role in enabling high-quality and immersive multimedia experiences that can better reflect the diversity and richness of human communication.

{\small
\bibliographystyle{ieee_fullname}
\bibliography{arxiv_paper.bib}
}

\end{document}